\documentclass[11pt]{article}

\usepackage[final]{acl}
\usepackage{tcolorbox}
\usepackage{adjustbox}
\usepackage{tabularx}
\usepackage{algorithm}
\usepackage{algpseudocode}
\usepackage{amsmath}

\usepackage{amssymb}
\usepackage{mathtools}
\usepackage{balance}
\usepackage{booktabs}
\usepackage{comment}
\usepackage{times}
\usepackage{latexsym}
\usepackage[T1]{fontenc}
\usepackage[utf8]{inputenc}
\usepackage{newunicodechar}
\usepackage{url} % для команды \url
\newunicodechar{≈}{\approx}

\usepackage{microtype}
\usepackage{titlesec} 
\titlespacing*{\section}{0pt}{*0.8}{*0.5} 
\titlespacing*{\subsection}{0pt}{*0.7}{*0.4} 
\usepackage{enumitem} 
\setlist[itemize]{noitemsep,topsep=0pt,leftmargin=*}
\setlist[enumerate]{noitemsep,topsep=0pt,leftmargin=*}

\usepackage{inconsolata}

\usepackage{graphicx}

\newcommand{\sysname}{CLARITY}
\newcommand{\smallcaption}{\captionsetup{font=small}}
\newcommand{\llmname}{GigaChat}

\title{\sysname: Clinical Assistant for Routing, Inference, and Triage}

\author{
\textbf{Vladimir Shaposhnikov}\textsuperscript{*,1,2}\quad
\textbf{Aleksandr Nesterov}\textsuperscript{*,1}\quad
\textbf{Ilia Kopanichuk}\textsuperscript{*,1,3}\quad \\
\textbf{Ivan Bakulin}\textsuperscript{1,3}\quad
\textbf{Egor Zhelvakov}\textsuperscript{1}\quad 
\textbf{Ruslan Abramov}\textsuperscript{4}\quad \\
\textbf{Ekaterina Tsapieva}\textsuperscript{5}\quad
\textbf{Iaroslav Bespalov}\textsuperscript{1}\quad
\textbf{Dmitry V. Dylov}\textsuperscript{1}\quad
\textbf{Ivan Oseledets}\textsuperscript{1}
\\[3pt]
\textsuperscript{1}AIRI \quad
\textsuperscript{2}Skoltech \quad
\textsuperscript{3}MIPT \quad
\textsuperscript{4}SberMedAI \quad
\textsuperscript{5}Sber \\
\textsuperscript{*} Equal contribution \\
Russia \\[3pt]
\texttt{\{shaposhnikov, nesterov, kopanichuk\}@airi.net}
}

\begin{document}
\maketitle
\begin{abstract}

We present \sysname{} (Clinical Assistant for Routing, Inference and Triage), an AI-driven platform designed to facilitate patient-to-specialist routing, clinical consultations, and severity assessment of patient conditions. Its hybrid architecture combines a Finite State Machine (FSM) for structured dialogue flows with collaborative agents that employ Large Language Model (LLM) to analyze symptoms and prioritize referrals to appropriate specialists. Built on a modular microservices framework, \sysname{} ensures safe, efficient, and robust performance, flexible and readily scalable to meet the demands of existing workflows and IT solutions in healthcare.

We report integration of our clinical assistant into a large-scale national interhospital platform, with more than 55,000 content-rich user dialogues completed within the two months of deployment, 2,500 of which were expert-annotated for subsequent validation. The validation results show that CLARITY surpasses human-level performance in terms of the first-attempt routing precision, naturally requiring up to 3 times shorter duration of the consultation than with a human. 

\end{abstract}

\maketitle

\section{Introduction}

The integration of LLMs into healthcare could truly transform the industry. These models demonstrate potential to improve diagnostic accuracy, optimize clinical workflows, and enhance the overall experience for patients and physicians.  
Dialog systems powered by LLMs are particularly promising, as they are capable of automating routine tasks, generating initial diagnostic hypotheses to optimize access to care and reduce waiting times \cite{amie_model, DISC_model, CoD_model, pubmed_llms_healthcare, pmc_llms_scoping_review}.

Medical dialogue systems are designed to facilitate clinical consultations, assist healthcare providers, and connect patients with proper expertise. Unlike general-purpose conversational AI, these systems must be able to handle complex diagnostic hypotheses reasoning, extended back-and-forth interactions, and domain-specific medical knowledge. Although recent advances in LLMs have significantly improved their conversational abilities, they are still fundamentally flawed for real-world adoption in healthcare.

Despite all the efforts, modern patient routing approaches often involve unnecessary delays. Patients typically consult general practitioners (GPs) as an intermediary step before being referred to specialized experts, which increases wait times and complicates timely access to care. Automated assessment of patient conditions with a generated diagnostic hypotheses could significantly streamline this process, benefiting both patients and healthcare providers \cite{umerenkovai, arxiv_llms_healthcare_review}.  This optimization also promises substantial economic advantages, potentially reducing healthcare costs and improving the allocation of resources.

The potential economic impact extends to the rapidly growing telemedicine market.  While telemedicine offers expanded access to care, high consultation costs remain a significant barrier. Automating certain GP functions through LLM-powered systems offers a pathway to more affordable and accessible telemedicine services \cite{telemed_review, telemed_primer}. Similarly, improvements in online appointment scheduling systems, often hindered by user interface challenges and integration issues, could further enhance patient experience and reduce administrative burdens \cite{scheduling_health1, scheduling_health2}.

Several specialized LLMs, including AMIE, DISC-MedLLM, CDD, and CoD \cite{amie_model, DISC_model, CoD_model}, demonstrate the potential of this technology.  While systems like AMIE show high diagnostic hypotheses accuracy, challenges remain regarding scalability and reliance on specialized datasets.  Others, such as DISC-MedLLM, prioritize empathetic patient interaction, while CDD focuses on structured symptom questioning.  However, the field lacks unified standards for training, evaluation, and deployment, hindering model comparison and regulatory approval \cite{jamanetwork_review, nature_medpalm}.

Despite these advancements, critical challenges persist.  LLMs may generate inaccurate or fabricated information ("hallucinations")\cite{hall_survey1, hall_survey2, hall_med_test}, a particularly dangerous risk in medicine.  Current models also struggle to reliably identify life-threatening conditions and may exhibit inconsistencies in patient interaction, potentially affecting trust.  These issues necessitate robust solutions to ensure safe and effective integration of LLMs into clinical practice \cite{paper_survey_basic, paper_survey_metrics, paper_survey_advanced}.

To address these challenges, this article introduces \sysname{}, a hybrid system integrating the strengths of rule-based and LLM approaches.  \sysname{} is designed to provide a flexible, controllable, and accurate medical dialog system suitable for real-world applications.  By leveraging FSM, a microservices architecture, and specialized datasets, \sysname{} facilitates structured, context-aware patient interactions, enhancing diagnostic hypotheses reliability, enabling real-time critical condition recognition, and providing personalized patient guidance.

\section{System Architecture}

The \sysname{} system is designed as a hybrid architecture, integrating an FSM for the management of dialogue with a microservices architecture for the processing of requests.

The architecture consists of the following core components:
\begin{itemize}
\item \textbf{FSM} governs the dialogue flow by managing states and transitions based on user input and context. %The FSM acts as the central orchestrator, coordinating interactions between different services.
\textbf{Text generation services} are responsible for a natural language generation and complex query processing. 
In our deployment, the LLM used for generation and analysis is \llmname{}\footnote{\url{https://giga.chat/}}.
 
\item \textbf{Decision making services} are responsible for input text classification and decision-making. %They provide boolean outputs that guide FSM transitions.
\item \textbf{Microservice architecture} ensures modularity and scalability by isolating individual components; real-time performance further depends on minimizing inter-service latency and optimizing orchestration. %This architecture allows for secure integration and independent updates of system modules.
\end{itemize}

\begin{figure}[t]
    \centering
    \smallcaption
    \includegraphics[width=.94\columnwidth]{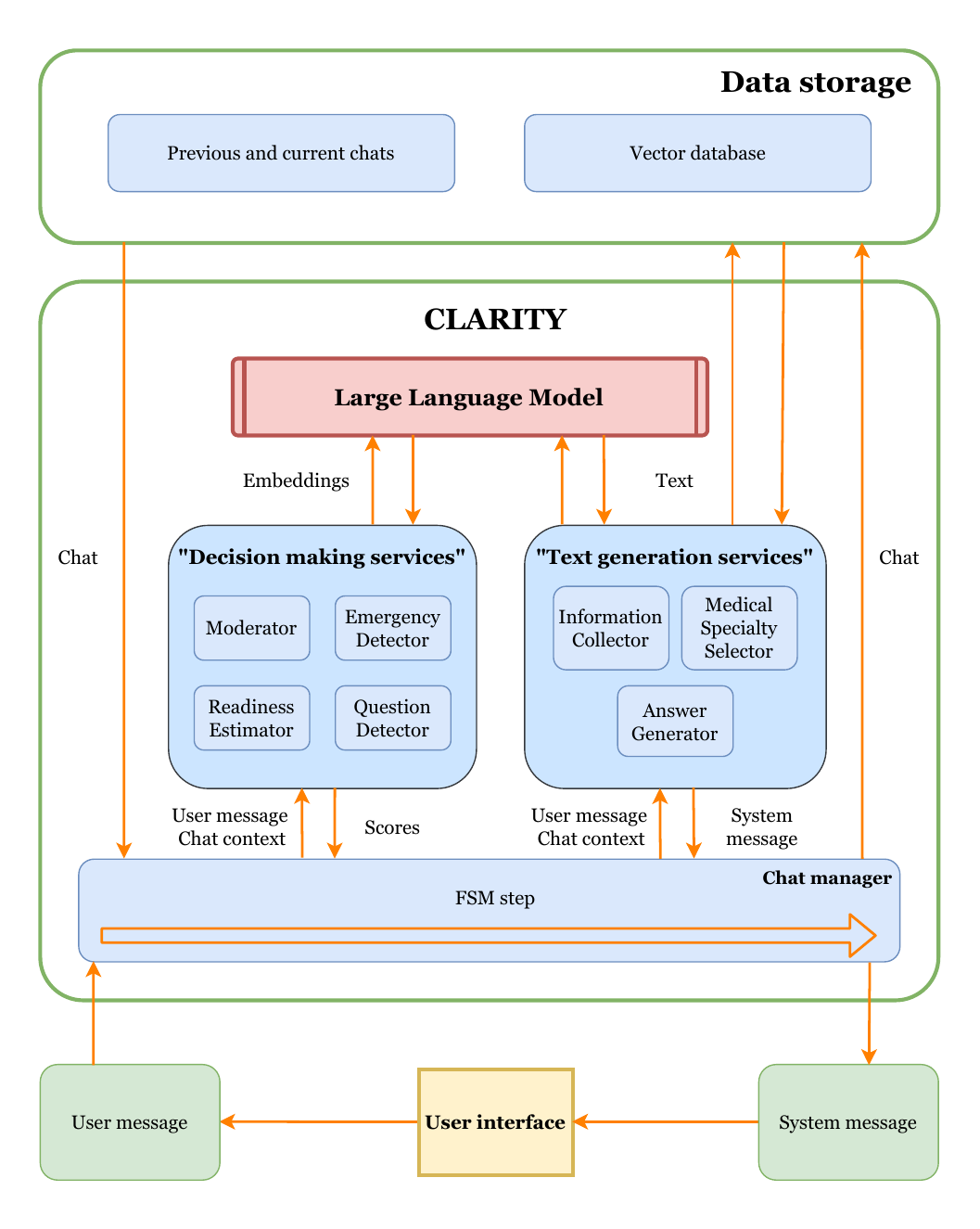}
    \caption{The architecture of the \sysname{} system.}
    \label{fig:system_architecture}

\end{figure}

\textbf{Figure~\ref{fig:system_architecture}} illustrates the architecture of the \sysname{} system, showing how its core components interact. The FSM serves as the backbone of dialogue management, ensuring structured and predictable interactions, while text generation and decision making services provide classification and response generation capabilities, respectively.

\subsection{Chat Manager}

The Chat Manager in the \sysname{} system is implemented as a FSM, tailored to specific requirements of the medical domain. Its primary goals are to ensure structured dialogue flow, maintain patient safety, and handle sequential information gathering.

\textbf{States.} The FSM has six dialogue contexts: \begin{itemize}
    \item \textit{Initialization} - greets the user and begins collecting initial complaints;
    \item \textit{Information Collection} - coordinates the information - gathering module for symptoms and medical history;
    \item \textit{Diagnostic Hypothesis \& Routing} - processes collected data to generate preliminary diagnostic hypotheses, select medical specialty, and formulate recommendations;
    \item \textit{Moderation} - prevents unsafe dialogue scenarios;
    \item \textit{Emergency} - executes predefined protocols for urgent cases;
    \item \textit{Free Dialogue} - handles non-standard requests via the answer-generation module.
\end{itemize} 

\textbf{Transitions.} Movement between states depends on user input, dialogue history, and contextual signals (see Appx.~\ref{appx:fsm_transition}); decisions are driven by outputs from dedicated decision-making services and the FSM’s internal logic.  

\textbf{Output Functions.} Each state defines its own system responses, encompassing natural-language generation by LLMs and calls to internal microservices (see Appx.~\ref{appx:fsm_outputs}).

For a more detailed formal description of the FSM and its components, please refer to Appendix \ref{appx:fsm_description}.

%To handle complex or ambiguous scenarios, the FSM employs an internal iterative cycle, allowing for sequential state transitions and service invocations. In scenarios where the standard dialog flow proves inadequate, the system transitions into a \textit{free dialogue state}, thereby invoking LLM-based modules to interact with the user.

By combining the structured control of the FSM with the adaptability of LLMs, the Chat Manager achieves a balance between \emph{reliability} and \emph{flexibility}. First, \textbf{predictability and safety} are ensured by the FSM's ability to control the dialogue flow and enforce strict transitions between states. Second, the FSM's modular design ensures \textbf{scalability}, making it easy to integrate new states, transitions, and services as needed. Finally, the FSM is specifically tailored to meet \textbf{medical domain requirements}. Moderation state, critical situation handling, and strict control mechanisms ensure the system adheres to the high safety and reliability standards demanded in this field.

\subsection{FSM Scalability}

CLARITY's FSM addresses scalability concerns through hierarchical decomposition designed to prevent architectural brittleness as clinical scope expands:

\textbf{Multi-level FSM Design.} The current implementation employs logical decomposition where a global FSM graph describes overall system behavior and manages high-level transitions between initialization, information collection, diagnostic hypotheses, moderation, and emergency handling states. Simultaneously, local FSM graphs govern the logic within individual medical scenarios and specialty-specific workflows, creating a hierarchical structure that naturally distributes complexity across multiple manageable components.

\textbf{Modular Expansion Strategy.} New medical domains can be integrated by: 
\begin{itemize}
    \item Defining domain-specific local FSM graphs that handle specialty workflows.
    \item Adding appropriate transition points in the global FSM to route users to new domains.
    \item Implementing corresponding microservices that provide domain-specific processing capabilities.
\end{itemize}
This approach ensures system complexity grows linearly rather than exponentially with clinical area additions.

\textbf{Graph-based Storage and Analysis.} The FSM is stored as a graph structure in memory, enabling development of semi-automated analysis tools for: redundant cycle detection, unreachable state identification, conflicting transition condition analysis, and other architectural consistency checks that support long-term maintainability.

\subsection{Services}
\subsubsection{Moderator}

The moderator checks whether a given request or model response contains a prohibited topic. A separate moderation skill is needed for two reasons. First, it is essential to limit sensitive topics, aggression, and hatred in conversations. Second, default moderation systems often cannot handle medical data due to compliance requirements. Moderation is formulated as a binary classification problem: 0 (no prohibited topics) or 1 (prohibited topics detected). The proposed algorithm uses a nested blacklist of prohibited words specific to each topic. Performance is evaluated using the $F_1$-score.
 
\subsubsection{Emergency Detector} 

The emergency detector checks whether a user is in need for emergency medical care. It is a binary classification problem: 0 (no emergency) or 1 (emergency). The algorithm processes input chat $W$ into a text string of user and model messages via concatenation. The function $\sigma$ is computed as:
\begin{equation}
\begin{aligned}
\sigma_{tr}(W) = \text{HGB}\Big(&\text{PCA}\big( \text{concat}(\text{tfidf}(W), \\
& \text{OHE}(W), \text{LLM}(W)), n_c \big) \Big) > t
\end{aligned}
\end{equation}

where $t$ is the threshold value, HGB is a hist-based gradient boosting, and $n_c = 423$ is the number of components for PCA. We enhance the concatenation function by including a one-hot-encoder (OHE) for critical words and a binary value indicating whether the LLM model considers the condition critical. Performance is evaluated using $F_1$-score and false positive rate (FPR).

\subsubsection{Readiness Estimator} %% Саша

The readiness estimator determines the appropriate time to transition from the information collection phase to the referral stage. It mitigates the risk of premature or delayed transitions, improving user experience and system accuracy. The module is trained on a corpus of 2,500 medical dialogues, using TF-IDF and contextual embeddings for feature extraction. Performance is evaluated using the mean absolute percentage error (MAPE) and the $F_1$-score.

\subsubsection{Question Detector} %% Саша

The question detector is responsible for the binary classification of user messages with the objective of determining whether a message constitutes a clarifying question. This module ensures flexibility in the system's workflow for collecting medical history and complaints. Performance is evaluated using $F_1$-score.

\subsubsection{Information Collector} %% Саша

The information collector module gathers comprehensive information on patient complaints and medical history. It includes a question generator based on a language model, a dialogue search system, and a question relevance classifier (see Appx. \ref{appx:InfCollDiag}). 

\begin{enumerate}
    \item The module begins by receiving the user’s message and dialogue history. 
    \item It then searches the historical dialogue vector database; if a similar case is found (cosine similarity $> 0.965$), previously generated questions are reused.
    \item If no match is identified, the system uses a domain-specific prompt to generate five new candidate questions with the LLM.
    \item To avoid redundancy, each candidate is compared to prior dialogue questions, and those with cosine similarity above 0.86 are discarded. 
    \item Remaining candidates are passed through a relevance classifier trained on 2,500 labeled examples; the classifier achieves 0.84 precision on the "relevant" class.
    \item The most relevant question is selected and presented to the user. This architecture ensures both contextual accuracy and low response latency.
\end{enumerate}

\begin{figure}[t]
\small
\begin{tcolorbox}[colback=white,colframe=black!20,boxrule=0.3pt,left=2pt,right=2pt,top=2pt,bottom=2pt]
\textbf{Doc:} What’s bothering you?\\
\textbf{Pt:} I have a headache.\\
\textbf{Doc:} Where exactly is the pain located?\\
\textbf{Pt:} The back of my head.\\
\textbf{Doc:} Are you experiencing any other symptoms, such as nausea or vomiting?\\
\textbf{Pt:} No.\\
\textbf{Doc:} How intense is the pain?\\
\textbf{Pt:} 5 out of 10.\\
\textbf{Doc:}\\
Cervicogenic headache – \emph{General practitioner}. Pain in the back of the head may be related to cervical‐spine issues.\\
Cervical osteochondrosis – \emph{Neurologist}. Neck pathology may provoke occipital pain.\\
\textbf{Doc:} Is everything clear? Feel free to ask questions!
\end{tcolorbox}
\vspace{-0.4em}
\caption{Dialogue example illustrating the \emph{Transparency} scenario.
Full dialogue examples for \emph{Critical}, \emph{Safety}, and
\emph{Adaptability} are provided in Appendix~\ref{appx:dialogue_examples}, Table~\ref{tab:example_full}.}
\label{fig:transparency_example}
\end{figure}

\subsubsection{Medical Specialty Selector} %% Володя 

The medical specialty selector is a core module that generates candidate diagnoses, identifies appropriate medical specialists, and provides supporting explanations. The process includes three sequential stages (Appx. \ref{appx:SpecSelDiagram}):

1. \textbf{Diagnostic Hypotheses Generation:} Using the patient’s complaints and medical history, the system queries the LLM to generate $N$ possible diagnostic hypotheses. If fewer than $N$ are returned, the request is repeated to ensure sufficient coverage.

2. \textbf{Specialist Selection:} For each diagnostic hypotheses, a parallel LLM call determines the most relevant medical specialist (e.g., cardiologist for cardiovascular symptoms, gastroenterologist for digestive issues), ensuring that each condition is matched with an appropriate expert.

3. \textbf{Explanation Generation:} The system produces a short explanation for each diagnosis-specialist pair, including a description of the condition, reasoning behind the diagnostic hypotheses, and rationale for the referral. These explanations aim to improve patient understanding and transparency.

The module is optimized via parallel processing and selective regeneration of incomplete outputs, which significantly reduces latency. Its performance is evaluated using pairwise precision and recall, comparing system recommendations to expert annotations.

\subsubsection{Answer Generator}

This module allows for the management of arbitrary user queries and open dialogues within the context of medical consultations. The module is activated in two scenarios:

1. A deviation from the dialogue script may occur when the user poses a question that falls outside the scope of the current predefined dialogue flow.

2. Once the standard dialogue script has been completed and all necessary information has been collected, as well as a preliminary diagnostic hypotheses provided, the dialogue can be considered complete. In this instance, the open-dialogue module permits the user to pose supplementary queries, seek elucidation, or request general health-related information.

The functionality of this module is based on the capabilities of LLMs, which are able to generate coherent and contextually relevant responses within a dialogue. In order to adapt the LLMs to the medical domain, a specialised system prompt is employed. This sets the context of the interaction, constrains the knowledge domain to medical topics and guides the model towards generating responses that are ethically sound and informative. This module facilitates smooth transitions between predefined dialogue scripts and open-ended discussions, ensuring the system's flexibility and its capacity to respond effectively to a diverse range of user queries.

\section{Experiments And Results}

%\subsection{Component-Level Evaluation}
In this section, we present a comprehensive evaluation of \sysname{}'s performance across its core components. The experiments were conducted on unique datasets, annotated by licensed doctors and medical experts to ensure high-quality ground truth labels. We evaluate the system using a combination of standard metrics (e.g., precision, recall, $F_1$-score) and custom metrics tailored to the medical domain (e.g., pairwise precision and recall for specialist selection). Additionally, we discuss the results of pilot studies, which demonstrate the system's practical applicability in real-world scenarios.

\begin{table}[t]
\centering
\footnotesize
\setlength{\tabcolsep}{3pt}
\renewcommand{\arraystretch}{0.9}
\caption{Services metrics}
\label{tab:services-metrics}
\begin{tabularx}{\linewidth}{lX}
\toprule
 & \textbf{Value} \\ \midrule
Transparency & Information collector (P = 84 \%, $f_{\text{rep}} = 0$); readiness estimator (MAPE = 22 \%); medical-specialty selector (R@3 = 96 \%)\\[2pt]
Critical     & Emergency detector (Precision = 72 \%, Recall = 49 \%, F$_1$ = 59 \%, FPR = 2 \%)\\[2pt]
Safety       & Moderator (F$_1$ = 95 \%)\\[2pt]
Adaptability & Question detector (F$_1$ = 94 \%)\\
\bottomrule
\end{tabularx}
\end{table}

\begin{table}[t]
\centering
\footnotesize
\setlength{\tabcolsep}{3pt}
\renewcommand{\arraystretch}{0.9}
\caption{Pilot statistics}
\label{tab:pilot-stats}
\begin{tabularx}{\linewidth}{lX}
\toprule
 & \textbf{Value} \\ \midrule
Transparency & 55 k+ dialogues; 3× faster than human; 80 \% “friendly”; 32.4 \% conversion\\[2pt]
Critical     & 7.4 \% dialogues flagged critical\\[2pt]
Safety       & 3.4 \% non-target dialogues\\[2pt]
Adaptability & 6.5 \% non-standard dialogues\\
\bottomrule
\end{tabularx}
\end{table}

Across the safety-critical modules, the \emph{Moderator} achieved an $F_1$-score of 0.95 with a false-positive rate below 1.5 \%, indicating that the system rarely blocks benign content while effectively filtering prohibited topics. The \emph{Emergency Detector} demonstrated a precision of 0.72, recall of 0.49, and an $F_1$-score of 0.59 at an operating point that keeps the false-positive rate below 0.02. This threshold balances the risk of missed emergencies against the clinical burden of false alarms. Together, these figures confirm the reliability of the safety layer that gates downstream actions.

The remaining components likewise show strong offline performance. The \emph{Readiness Estimator} attains an $F_1$ of 0.78 and a mean absolute percentage error of 22 \% when forecasting the anamnesis phase, allowing the dialogue manager to time interventions accurately. The \emph{Question Detector} reaches macro-$F_1=0.94$ (recall$_\text{pos}=0.87$, precision$_\text{neg}=0.99$), ensuring that clarification prompts are issued only when needed. The \emph{Information Collector} maintains 0.84 precision for question relevance while responding in 5 s on average, with 20 \% of prompts served from cache. Finally, the \emph{Medical-Specialty Selector} ranks its recommendations effectively: its top suggestion is correct in 80 \% of cases, and the top two cover 95 \%, cor-responding to Precision@1 = 77 \% and Recall@3 = 96 \%. These results collectively demonstrate that each subsystem meets the accuracy and latency requirements necessary for safe real-world deployment.

A comprehensive breakdown of datasets, model architectures and training protocols is available in Appendix~\ref{appx:detailed_results}, while Tables \ref{tab:services-metrics} and \ref{tab:pilot-stats} collate the key service-level metrics and pilot statistics across all four scenarios. Figure \ref{fig:transparency_example} reproduces a complete \emph{Transparency} consultation, full transcripts for the remaining cases - \emph{Critical}, \emph{Safety}, and \emph{Adaptability} - are available in Appendix \ref{appx:dialogue_examples} (Table \ref{tab:example_full}).

\subsection{First Pilot Study: Understanding User Behavior And Preferences}
A first pilot study was conducted at the end of the fourth quarter of 2024 on an active dialogue platform with the aim of evaluating the effectiveness of \sysname{} in real-world conditions, as well as identifying user preferences and potential avenues for improvement. The study encompassed a comprehensive assessment of socio-demographic indicators and a detailed analysis of user preferences. Moreover, the pilot incorporated an in-depth examination of user feedback on the current functionality of the system, offering invaluable insights into the usability and user experience.

The pilot studies were conducted on a nationwide inter-hospital telehealth platform where \sysname{} was fully integrated into the standard user onboarding and consultation workflow; users interacted with the system in their usual telehealth environment (both new registrants and returning users). Participants received clear onboarding notifications about the AI assistant and could request a hand-off to a human specialist at any time. %further details on consent and onboarding procedures are provided in Appendix.

A total of 64\% of users who completed the onboarding process for the services in question proceeded to initiate a dialogue. Of the 1,534 initiated consultations, 62.13\% resulted in recommendations from a specialist and a diagnostic hypothesis. The mean time taken for an appointment to conclude, including the provision of recommendations, was 2 minutes 13 seconds. This is considerably less than the time typically allocated for an in-person consultation with a general medical practitioner.

The pilot study resulted in 785 distinct preliminary diagnostic hypotheses, with referrals made to 70 physicians across various specialties. The most frequently recommended specialists were neurologists, otolaryngologists, and gastroenterologists, which is indicative of the relevance of these specialties within the scope of the pilot study (Appendix \ref{appx:recom_spec}). Neurological, renal, and gastrointestinal disorders were the most commonly diagnosed conditions. A comparison of the socio-demographic characteristics of the participants and those of the overall user base of the platform indicated a correspondence between the two groups in terms of both age and gender (Appendix \ref{appx:gender_dist} and Appendix \ref{appx:age_dist}).

A total of 18.2\% of users indicated that they found the consultation to be fully useful, while 54.5\% of respondents rated it as partially useful. However, 31\% of users reported that the number of questions asked was insufficient, potentially pointing to limitations in audience targeting or the depth of the dialogue process. On a positive note, 80\% of users rated the interaction as friendly, which is a key factor in maintaining user engagement and satisfaction (Appendix \ref{appx:commun_friendly}).

\subsection{Second Pilot Study: Focus On Real-World Application}
A second pilot study was conducted to evaluate the real-world application of \sysname{} in a larger-scale setting. This study focused less on audience behavior and more on the practical implementation and outcomes of the system. During the pilot period, a total of 55{,}856 dialogues were conducted. The dialogue funnel analysis revealed a 92.3\% conversion rate to the next step after dialogue initiation, with 54.7\% of dialogues resulting in diagnostic hypotheses and a referral to a physician. In all cases, \sysname{} performed the initial triage and routed users toward the appropriate channel. For appointment conversions, “online” consultations (telemedicine appointments booked and conducted via the platform) converted at ~26.8\%, whereas “offline” consultations (in-person visits at clinics or hospitals) converted at ~5.6\%.

The second pilot study demonstrated the scalability and practical utility of \sysname{} in real-world conditions, with a high rate of dialogue progression and a significant proportion of cases leading to actionable medical outcomes. The conversion rates to both online and offline consultations further underscore the system's effectiveness in facilitating user engagement and follow-through.

\section{Discussion} %% Ярослав

The \sysname{} system demonstrates the effectiveness of a hybrid approach to medical dialogue systems, combining structured FSM-based dialogue management with the flexibility of large language models. Our experimental results validate this architecture's advantages while revealing important insights about AI deployment in healthcare settings.

\paragraph{Clinical Accuracy and Routing Efficiency.}
The system achieved remarkable performance in specialist routing, with 77\text{\%} precision for first recommendations and 96\text{\%} recall for top-3 recommendations. This surpasses typical human-level performance in initial routing while requiring only one-third of the usual consultation time  (mean 2m 13s vs $\approx$ 6-7 minutes for GP practice).
These results are particularly significant given current healthcare bottlenecks. The traditional model, where patients must first consult general practitioners for specialist referrals, creates substantial delays in accessing specialized care. \sysname{} demonstrates that AI systems can effectively assume this initial triage function, potentially accelerating patient access to appropriate specialists while reducing the burden on primary care providers.

\paragraph{Critical Condition Recognition}
The emergency detection figures confirm an \textbf{operationally acceptable
performance} under the current stakeholder-defined false-alarm budget, while highlighting the need for future recall improvements. In real-world deployment, the system identified 7.4\text{\%} of cases as requiring urgent intervention, aligning with emergency department triage statistics. This achievement is particularly noteworthy as existing medical AI systems often struggle with reliable critical condition identification. \sysname{}'s hybrid architecture, combining FSM rules with LLM flexibility, provides a more dependable approach to detecting potentially dangerous situations.

%\paragraph{Real-World Performance and User Interaction}
%The extensive pilot deployment of over 55,000 dialogues revealed several key findings about the system's practical viability.

\paragraph{Engagement Metrics.} The 32.4\text{\%} conversion rate to actual appointments indicates strong user trust and clinical relevance, substantially higher than typical digital platform metrics.

\paragraph{User Experience.} The 80\text{\%} positive interaction ratings demonstrate the system's ability to maintain empathetic and professional communication while adhering to clinical protocols. Operational Efficiency: The mean consultation time of 2 minutes 13 seconds represents a significant improvement over traditional consultations, without compromising diagnostic hypotheses accuracy.
\paragraph{Dialogue Control.} The readiness estimator's F1-score of 78\text{\%} and the question detector's accuracy of 94\text{\%} demonstrate \sysname{}'s sophisticated dialogue management capabilities. The near-zero repetition rate (frep = 0\text{\%}) in information collection confirms efficient information gathering without redundancy. Only 3.4\text{\%} non-target dialogues indicates exceptional conversation management, addressing a common limitation of pure LLM-based systems.

%\paragraph{Dialogue Management Innovation}
% The readiness estimator's F1-score of 78\text{\%} and the question detector's accuracy of 94\text{\%} demonstrate CLARITY's sophisticated dialogue management capabilities. The near-zero repetition rate (frep = 0\text{\%}) in information collection confirms efficient information gathering without redundancy. These metrics meet our goal of addressing fundamental challenges in medical dialogue systems.
% Unlike existing solutions based solely on LLMs, our hybrid architecture provides several distinct advantages:\\
% \textbf{Structured Reasoning:} The FSM ensures consistent diagnostic pathways while allowing for dynamic adaptation to patient responses.\\
% \textbf{Safety Assurance:} The multi-layer architecture with specialized components for moderation and emergency detection provides robust safety guarantees.\\
% \textbf{Scalability:} The microservices architecture enables seamless integration with existing healthcare systems while maintaining performance at scale.

% The integration of retrieval-augmented generation (RAG) further enhances system reliability by grounding responses in verified medical knowledge, addressing the critical challenge of hallucination in medical AI applications. This approach achieved a 94\text{\%} accuracy rate in maintaining factual consistency, significantly higher than the baseline LLM performance.

These results suggest that successful medical AI systems require more than just powerful language models -- they call for carefully designed hybrid pipelines with the strengths of different approaches of strict safety and reliability standards. \sysname{}'s performance metrics demonstrate that such hybrid systems can achieve both the high accuracy and the appreciation of users, creating a viable path for AI integration in a clinical setting.

\section{Conclusion} %% Ярослав

In conclusion, \sysname{} advances the field of medical dialogue systems by successfully addressing the major challenges that have hindered the widespread adoption of AI-powered conversational agents in healthcare. By seamlessly integrating structured reasoning, multi-agent collaboration, and robust safety measures, \sysname{} sets a new standard for reliable, efficient, and user-centric medical assistance.

The system's strong performance in real-world settings, coupled with the positive user feedback, showcases its potential to facilitate the delivery of healthcare services.
The insights gained from the development and evaluation of \sysname{} will undoubtedly inspire and guide future efforts in this field, ultimately leading to the development of more accessible, efficient, and personalized healthcare assistants.

\section{Contribution Statement}
\textbf{Vladimir Shaposhnikov}, \textbf{Aleksandr Nesterov}, and \textbf{Ilia Kopanichuk} contributed equally: they formulated the research, designed and ran the experiments, performed the analysis, and drafted the core manuscript text. \textbf{Ekaterina Tsapieva} and \textbf{Ruslan Abramov} led the productization effort and guided the introduction of the solution into a real-world workflow. \textbf{Ivan Bakulin} and \textbf{Egor Zhelvakov} were responsible for technical deployment and production integration. \textbf{Iaroslav Bespalov}, \textbf{Dmitry V.~Dylov} and \textbf{Ivan Oseledets} are co-senior authors who supervised the project and analysed the results. All authors took part in preparing the final manuscript.

\section{Limitations}
Despite the strengths of the \sysname{} system, several limitations should be acknowledged:
\begin{itemize}
    \item \textbf{Limited interpretability of LLM outputs.} While the system provides structured responses and intermediate explanations, it does not offer full introspection into model reasoning. Improving interpretability is an ongoing research challenge in medical AI. In future iterations, we plan to incorporate chain-of-thought tracing and example-based rationales to make the decision process more transparent for clinical users.
    
    \item \textbf{Model identity disclosure.} The system uses \llmname{} as the foundation model for generation and analysis. 
    While this work is designed to be LLM-agnostic, all reported results were obtained with \llmname{} under the production configuration used in deployment.

    \item \textbf{Generalization to rare or atypical conditions.} The system has primarily been validated on common and well-represented clinical cases, as these reflect the majority of real-world usage scenarios where \sysname{} is currently deployed. Evaluation on rare diseases, complex multi-morbidity cases, and ambiguous complaints remains limited. We plan to address this in future development by expanding the training dataset and annotation pipeline to include synthetic and curated edge cases, thereby supporting broader and more robust generalization.

    \item \textbf{Dialogue depth and coverage.} In pilot studies, 31\% of users reported that the consultation asked “too few” questions. A formal root-cause analysis has not yet been performed; it is scheduled for Q4 2025 and will be led by an interdisciplinary team of clinicians and product specialists. The study will review a stratified sample of under-questioned dialogs to identify failure modes (e.g.\ premature readiness cut-off or vague initial complaints) and will feed its findings into the next iteration of the Readiness Estimator and Information Collector. Planned mitigations include (i) adding dialogue-length and answer-entropy features to the estimator and (ii) augmenting the training cache with synthetic edge-case dialogs. Our target is to lift the “question sufficiency” satisfaction rate above 85\% while keeping the average consultation time unchanged.

    \item  \textbf{Time–Satisfaction Mismatch and Explanation Factuality Limit} We acknowledge that shorter consultations do not necessarily translate into higher user satisfaction or trust: although median time decreased, many users still rated interactions as only partially useful, and we did not establish a causal link between duration, perceived usefulness, and trust. In addition, while we evaluated routing accuracy, we did not conduct a dedicated factuality (hallucination) audit of generated explanations. Future iterations will include targeted user-trust studies and clinician-in-the-loop audits of explanation correctness.

    \item \textbf{Lack of clinical outcome validation.} Although diagnostic hypotheses and referral suggestions were quantitatively evaluated, long-term patient outcomes were not tracked. This is due to regulatory and data privacy limitations in pilot deployments. As part of ongoing collaborations with clinical partners, we plan to conduct retrospective chart reviews and, where possible, prospective outcome-linked evaluations in future studies.

    \item \textbf{Absence of active learning pipeline.} Currently, CLARITY does not implement real-time self-updating mechanism or active learning. This design choice reflects critical regulatory and ethical constraints guiding the deployment of medical AI. Medical AI systems must pass stringent validation before algorithmic changes can be implemented, making real-time updates impossible from a regulatory standpoint. Furthermore, self-updating models based on patient interactions raise privacy concerns and may lead to unpredictable behavioural changes that undermine confidence in clinical decision-making. As future work, we are developing a controlled active learning pipeline that will operate under strict supervision and include offline batch learning cycles with mandatory clinical validation, explicit patient consent for the use of anonymised data, and differential privacy techniques.

    \item \textbf{Triage performance metrics.} The emergency-detection module is currently operated at a threshold chosen by clinical stakeholders to keep the false-positive rate below 0.02, because every alert triggers an on-call physician review and may result in ambulance dispatch. This conservative setting yields precision 0.72, recall 0.49 and F$_1$ = 0.59 on the expert-annotated test set. \textit{Operational rationale:} in our deployment scenario($\approx$ 7\% of dialogues genuinely critical) a higher recall would double the number of alerts and overwhelm human triage resources, whereas missed cases remain subject to the platform’s standard “red-flag” questionnaire shown to all users.
    
    \textit{Planned mitigation.}  A two-stage cascade is under development: (i) a soft threshold $t{=}0.05$ to maximise sensitivity, followed by (ii) an LLM self-consistency vote (three parallel prompts, majority $\ge 2/3$) that filters obvious false alarms.  The final choice of operating point will depend on a governance-board review of workload implications once additional deployment data are available.  All prospective changes will be audited against the same expert-annotated test set to ensure that patient safety is never traded for convenience.

    \item \textbf{Partial reproducibility.} We are committed to open-sourcing all available components of our system. The code will be publicly released at: \url{https://github.com/AnonymousSubmission996/CLARITY} upon acceptance. Certain production-specific services remain under NDA and cannot be distributed.
    
    \item \textbf{Lack of unified quantitative benchmarks.} There is still no publicly available dataset or protocol that simultaneously covers \emph{triage}, \emph{speciality selection}, \emph{emergency detection}, and \emph{safety filtering}.  Numerical head-to-head evaluation with earlier systems is therefore impossible without re-implementing each approach on private clinical data. As a partial remedy we performed a \emph{qualitative, functional comparison} of CLARITY with seven representative systems (AMIE, DISC-MedLLM, CoD, Polaris, MedAgents, Dual-Inf, and other).  The resulting matrix - Appendix~\ref{appx:related_works} Table~\ref{tab:medical_ai_comparison} - contrasts core capabilities and highlights \sysname{}’s unique FSM-based architecture.  While this analysis cannot replace a common benchmark, it offers readers a transparent, like-for-like view of where our system stands relative to prior art and exposes gaps that future community datasets should address.
    
    \textbf{Next step.}  We will publish an open evaluation \emph{protocol} (labeling guidelines, task definitions, scoring scripts) that lets any third party apply our four-task suite to \emph{their own} data without sharing protected records. This avoids legal constraints on commercial medical logs while enabling reproducible, apples-to-apples benchmarking across independent institutions.
    
\end{itemize}

\section{Ethics}

\paragraph{Regulatory compliance.}
Under the applicable national data-protection legislation, studies that use \emph{fully de-identified} service logs collected with \emph{explicit informed consent} are classified as non-interventional quality-improvement research and therefore \textbf{exempt from mandatory Institutional Review Board review}. All analyses were conducted on anonymised records stored in an encrypted, access-controlled environment.

\paragraph{Informed consent and data handling.}
All users of the telehealth service explicitly agreed that their anonymised dialogue records may be used for research and quality-improvement purposes.  Personally identifiable information (names, free-text IDs, contact details, timestamps finer than one hour) is removed at ingestion by an automated de-identification pipeline.  Only the resulting anonymised logs were accessible to the research team.

\paragraph{Model safety and audit.}
CLARITY enforces a three-layer safety framework: (i) a domain-specific moderation filter, (ii) conservative emergency-detection thresholds, and (iii) deterministic FSM
transitions that bound interaction depth.  All prompts and model outputs are logged and periodically audited by licensed physicians.

\section{Acknowledgments}
We are grateful to \textbf{SberHealth} for supporting the project and to all participating physicians for their invaluable feedback during development and evaluation. We also thank our clinical partners and annotators for their assistance with data labeling, which was essential for validating and improving the study's outcomes.

\balance
\bibliography{literature}

\begin{thebibliography}{31}
\providecommand{\natexlab}[1]{#1}

\bibitem[{Ala et~al.(2023)Ala, Simic, Deveci, and Pamucar}]{scheduling_health1}
Ali Ala, Vladimir Simic, Muhammet Deveci, and Dragan Pamucar. 2023.
\newblock \href {https://doi.org/10.1007/s11831-022-09855-z} {Simulation-based analysis of appointment scheduling system in healthcare services: A critical review}.
\newblock \emph{Archives of Computational Methods in Engineering}, 30(3):1961--1978.

\bibitem[{Bao et~al.(2023)Bao, Chen, Xiao, Ren, Wu, Zhong, Peng, Huang, and Wei}]{DISC_model}
Zhijie Bao, Wei Chen, Shengze Xiao, Kuang Ren, Jiaao Wu, Cheng Zhong, Jiajie Peng, Xuanjing Huang, and Zhongyu Wei. 2023.
\newblock \href {https://arxiv.org/abs/2308.14346} {Disc-medllm: Bridging general large language models and real-world medical consultation}.
\newblock \emph{Preprint}, arXiv:2308.14346.

\bibitem[{Barnett et~al.(1987)Barnett, Cimino, Hupp, and Hoffer}]{DXplain}
G~O Barnett, J~J Cimino, J~A Hupp, and E~P Hoffer. 1987.
\newblock {DXplain}. an evolving diagnostic decision-support system.
\newblock \emph{JAMA}, 258(1):67--74.

\bibitem[{Bedi et~al.(2024)Bedi, Liu, Orr-Ewing, Dash, Koyejo, Callahan, Fries, Wornow, Swaminathan, Lehmann, Hong, Kashyap, Chaurasia, Shah, Singh, Tazbaz, Milstein, Pfeffer, and Shah}]{jamanetwork_review}
Suhana Bedi, Yutong Liu, Lucy Orr-Ewing, Dev Dash, Sanmi Koyejo, Alison Callahan, Jason~A Fries, Michael Wornow, Akshay Swaminathan, Lisa~Soleymani Lehmann, Hyo~Jung Hong, Mehr Kashyap, Akash~R Chaurasia, Nirav~R Shah, Karandeep Singh, Troy Tazbaz, Arnold Milstein, Michael~A Pfeffer, and Nigam~H Shah. 2024.
\newblock Testing and evaluation of health care applications of large language models: A systematic review.
\newblock \emph{JAMA}.

\bibitem[{Chen et~al.(2024)Chen, Gui, Gao, Ji, Wang, Wan, and Wang}]{CoD_model}
Junying Chen, Chi Gui, Anningzhe Gao, Ke~Ji, Xidong Wang, Xiang Wan, and Benyou Wang. 2024.
\newblock \href {https://arxiv.org/abs/2407.13301} {Cod, towards an interpretable medical agent using chain of diagnosis}.
\newblock \emph{Preprint}, arXiv:2407.13301.

\bibitem[{Downes et~al.(2015)Downes, Mervin, Byrnes, and Scuffham}]{telemed_review}
Martin~J. Downes, Merehau~C. Mervin, Joshua~M. Byrnes, and Paul~A. Scuffham. 2015.
\newblock \href {https://doi.org/10.1186/s13643-015-0115-2} {Telemedicine for general practice: a systematic review protocol}.
\newblock \emph{Systematic Reviews}, 4(1):134.

\bibitem[{Gupta and Denton(2008)}]{scheduling_health2}
Diwakar Gupta and Brian Denton. 2008.
\newblock \href {https://doi.org/10.1080/07408170802165880} {Appointment scheduling in health care: Challenges and opportunities}.
\newblock \emph{IIE Transactions}, 40:800--819.

\bibitem[{Huang et~al.(2024)Huang, Yu, Ma, Zhong, Feng, Wang, Chen, Peng, Feng, Qin, and Liu}]{hall_survey1}
Lei Huang, Weijiang Yu, Weitao Ma, Weihong Zhong, Zhangyin Feng, Haotian Wang, Qianglong Chen, Weihua Peng, Xiaocheng Feng, Bing Qin, and Ting Liu. 2024.
\newblock \href {https://doi.org/10.1145/3703155} {A survey on hallucination in large language models: Principles, taxonomy, challenges, and open questions}.
\newblock \emph{ACM Trans. Inf. Syst.}
\newblock Just Accepted.

\bibitem[{Kühnel et~al.(2023)Kühnel, Jovanović, Hoffman, Golde, Schneider, and Hirsch}]{adahealth}
S~Kühnel, M.~Jovanović, H.~Hoffman, S.~Golde, L.~Schneider, and M.~Hirsch. 2023.
\newblock \href {https://doi.org/10.6084/m9.figshare.22250365.v1} {{Introduction of a pathophysiology-based diagnostic decision support system (DDSS) and its potential impact on the use of AI in healthcare}}.

\bibitem[{Loukachevitch et~al.(2023)Loukachevitch, Manandhar, Baral, Rozhkov, Braslavski, Ivanov, Batura, and Tutubalina}]{NERELBIO}
Natalia Loukachevitch, Suresh Manandhar, Elina Baral, Igor Rozhkov, Pavel Braslavski, Vladimir Ivanov, Tatiana Batura, and Elena Tutubalina. 2023.
\newblock \href {https://doi.org/10.1093/bioinformatics/btad161} {{NEREL-BIO: A Dataset of Biomedical Abstracts Annotated with Nested Named Entities}}.
\newblock \emph{Bioinformatics}.
\newblock Btad161.

\bibitem[{Loukachevitch et~al.(2024)Loukachevitch, Sakhovskiy, and Tutubalina}]{loukachevitch2024biomedical}
Natalia Loukachevitch, Andrey Sakhovskiy, and Elena Tutubalina. 2024.
\newblock Biomedical concept normalization over nested entities with partial umls terminology in russian.
\newblock In \emph{Proceedings of the 2024 Joint International Conference on Computational Linguistics, Language Resources and Evaluation (LREC-COLING 2024)}, pages 2383--2389.

\bibitem[{Meng et~al.(2024)Meng, Yan, Zhang, Liu, Cui, Yang, Zhang, Cao, Wang, Wang, Gao, Wang, Ji, Qiu, Li, Qian, Guo, Ma, Wang, Guo, Lei, Shao, Wang, Fan, and Tang}]{pmc_llms_scoping_review}
Xiangbin Meng, Xiangyu Yan, Kuo Zhang, Da~Liu, Xiaojuan Cui, Yaodong Yang, Muhan Zhang, Chunxia Cao, Jingjia Wang, Xuliang Wang, Jun Gao, Yuan-Geng-Shuo Wang, Jia-Ming Ji, Zifeng Qiu, Muzi Li, Cheng Qian, Tianze Guo, Shuangquan Ma, Zeying Wang, and 6 others. 2024.
\newblock The application of large language models in medicine: A scoping review.
\newblock \emph{iScience}, 27(5):109713.

\bibitem[{Moramarco et~al.(2021)Moramarco, Korfiatis, Savkov, and Reiter}]{babylong}
Francesco Moramarco, Alex~Papadopoulos Korfiatis, Aleksandar Savkov, and Ehud Reiter. 2021.
\newblock \href {https://arxiv.org/abs/2104.04402} {A preliminary study on evaluating consultation notes with post-editing}.
\newblock \emph{Preprint}, arXiv:2104.04402.

\bibitem[{Mukherjee et~al.(2024)Mukherjee, Gamble, Ausin, Kant, Aggarwal, Manjunath, Datta, Liu, Ding, Busacca, Bianco, Sharma, Lasko, Voisard, Harneja, Filippova, Meixiong, Cha, Youssefi, Buvanesh, Weingram, Bierman-Lytle, Mangat, Parikh, Godil, and Miller}]{polaris}
Subhabrata Mukherjee, Paul Gamble, Markel~Sanz Ausin, Neel Kant, Kriti Aggarwal, Neha Manjunath, Debajyoti Datta, Zhengliang Liu, Jiayuan Ding, Sophia Busacca, Cezanne Bianco, Swapnil Sharma, Rae Lasko, Michelle Voisard, Sanchay Harneja, Darya Filippova, Gerry Meixiong, Kevin Cha, Amir Youssefi, and 7 others. 2024.
\newblock \href {https://arxiv.org/abs/2403.13313} {Polaris: A safety-focused llm constellation architecture for healthcare}.
\newblock \emph{Preprint}, arXiv:2403.13313.

\bibitem[{Nazi and Peng(2024)}]{arxiv_llms_healthcare_review}
Zabir~Al Nazi and Wei Peng. 2024.
\newblock Large language models in healthcare and medical domain: A review.
\newblock In \emph{Informatics}, volume~11, page~57. MDPI.

\bibitem[{Nesterov et~al.(2025)Nesterov, Sakhovskiy, Sviridov, Valiev, Makharev, Anokhin, Zubkova, and Tutubalina}]{nesterov2025ruccod}
Aleksandr Nesterov, Andrey Sakhovskiy, Ivan Sviridov, Airat Valiev, Vladimir Makharev, Petr Anokhin, Galina Zubkova, and Elena Tutubalina. 2025.
\newblock Ruccod: Towards automated icd coding in russian.

\bibitem[{Nesterov et~al.(2022)Nesterov, Zubkova, Miftahutdinov, Kokh, Tutubalina, Shelmanov, Alekseev, Avetisian, Chertok, and Nikolenko}]{Nesterov2022239}
Alexandr Nesterov, Galina Zubkova, Zulfat Miftahutdinov, Vladimir Kokh, Elena Tutubalina, Artem Shelmanov, Anton Alekseev, Manvel Avetisian, Andrey Chertok, and Sergey Nikolenko. 2022.
\newblock \href {https://doi.org/10.18653/v1/2022.findings-acl.21} {{R}u{CC}o{N}: Clinical concept normalization in {R}ussian}.
\newblock pages 239--245.

\bibitem[{Pal et~al.(2023)Pal, Umapathi, and Sankarasubbu}]{hall_med_test}
Ankit Pal, Logesh~Kumar Umapathi, and Malaikannan Sankarasubbu. 2023.
\newblock \href {https://arxiv.org/abs/2307.15343} {Med-halt: Medical domain hallucination test for large language models}.
\newblock \emph{Preprint}, arXiv:2307.15343.

\bibitem[{Shi et~al.(2024)Shi, Liu, Du, Wang, Wang, Guo, Ruan, Xu, and Zhang}]{paper_survey_advanced}
Xiaoming Shi, Zeming Liu, Li~Du, Yuxuan Wang, Hongru Wang, Yuhang Guo, Tong Ruan, Jie Xu, and Shaoting Zhang. 2024.
\newblock \href {https://arxiv.org/abs/2405.10630} {Medical dialogue: A survey of categories, methods, evaluation and challenges}.
\newblock \emph{Preprint}, arXiv:2405.10630.

\bibitem[{Shortliffe(1977)}]{mycin}
E~H Shortliffe. 1977.
\newblock Mycin: A {Knowledge-Based} computer program applied to infectious diseases.
\newblock \emph{Proc Annu Symp Comput Appl Med Care}, pages 66--69.

\bibitem[{Singhal et~al.(2023)Singhal, Azizi, Tu, Mahdavi, Wei, Chung, Scales, Tanwani, Cole-Lewis, Pfohl, Payne, Seneviratne, Gamble, Kelly, Babiker, Sch{\"a}rli, Chowdhery, Mansfield, Demner-Fushman, Ag{\"u}era Y~Arcas, Webster, Corrado, Matias, Chou, Gottweis, Tomasev, Liu, Rajkomar, Barral, Semturs, Karthikesalingam, and Natarajan}]{nature_medpalm}
Karan Singhal, Shekoofeh Azizi, Tao Tu, S~Sara Mahdavi, Jason Wei, Hyung~Won Chung, Nathan Scales, Ajay Tanwani, Heather Cole-Lewis, Stephen Pfohl, Perry Payne, Martin Seneviratne, Paul Gamble, Chris Kelly, Abubakr Babiker, Nathanael Sch{\"a}rli, Aakanksha Chowdhery, Philip Mansfield, Dina Demner-Fushman, and 13 others. 2023.
\newblock Large language models encode clinical knowledge.
\newblock \emph{Nature}, 620(7972):172--180.

\bibitem[{Tang et~al.(2024)Tang, Zou, Zhang, Li, Zhao, Zhang, Cohan, and Gerstein}]{MedAgents}
Xiangru Tang, Anni Zou, Zhuosheng Zhang, Ziming Li, Yilun Zhao, Xingyao Zhang, Arman Cohan, and Mark Gerstein. 2024.
\newblock \href {https://arxiv.org/abs/2311.10537} {Medagents: Large language models as collaborators for zero-shot medical reasoning}.
\newblock \emph{Preprint}, arXiv:2311.10537.

\bibitem[{Tonmoy et~al.(2024)Tonmoy, Zaman, Jain, Rani, Rawte, Chadha, and Das}]{hall_survey2}
S.~M Towhidul~Islam Tonmoy, S~M~Mehedi Zaman, Vinija Jain, Anku Rani, Vipula Rawte, Aman Chadha, and Amitava Das. 2024.
\newblock \href {https://arxiv.org/abs/2401.01313} {A comprehensive survey of hallucination mitigation techniques in large language models}.
\newblock \emph{Preprint}, arXiv:2401.01313.

\bibitem[{Tripathi et~al.(2024)Tripathi, Sukumaran, and Cook}]{pubmed_llms_healthcare}
Satvik Tripathi, Rithvik Sukumaran, and Tessa~S Cook. 2024.
\newblock Efficient healthcare with large language models: optimizing clinical workflow and enhancing patient care.
\newblock \emph{J. Am. Med. Inform. Assoc.}, 31(6):1436--1440.

\bibitem[{Tu et~al.(2024)Tu, Palepu, Schaekermann, Saab, Freyberg, Tanno, Wang, Li, Amin, Tomasev, Azizi, Singhal, Cheng, Hou, Webson, Kulkarni, Mahdavi, Semturs, Gottweis, Barral, Chou, Corrado, Matias, Karthikesalingam, and Natarajan}]{amie_model}
Tao Tu, Anil Palepu, Mike Schaekermann, Khaled Saab, Jan Freyberg, Ryutaro Tanno, Amy Wang, Brenna Li, Mohamed Amin, Nenad Tomasev, Shekoofeh Azizi, Karan Singhal, Yong Cheng, Le~Hou, Albert Webson, Kavita Kulkarni, S~Sara Mahdavi, Christopher Semturs, Juraj Gottweis, and 6 others. 2024.
\newblock \href {https://arxiv.org/abs/2401.05654} {Towards conversational diagnostic ai}.
\newblock \emph{Preprint}, arXiv:2401.05654.

\bibitem[{Umerenkov et~al.(2025)Umerenkov, Nesterov, Shaposhnikov, Abramov, Romanenko, Kokh, Kirina, Abrosimov, Dylov, and Oseledets}]{umerenkovai}
Dmitriy Umerenkov, Alexandr Nesterov, Vladimir Shaposhnikov, Ruslan Abramov, Nikolay Romanenko, Vladimir Kokh, Marina Kirina, Anton Abrosimov, Dmitry Dylov, and Ivan Oseledets. 2025.
\newblock \href {https://doi.org/10.24963/ijcai.2025/1098} {Ai diagnostic assistant (aida): A predictive model for diagnoses from health records in clinical decision support systems}.
\newblock pages 9880--9889.

\bibitem[{Waller and Stotler(2018)}]{telemed_primer}
Morgan Waller and Chad Stotler. 2018.
\newblock \href {https://doi.org/10.1007/s11882-018-0808-4} {Telemedicine: a primer}.
\newblock \emph{Current Allergy and Asthma Reports}, 18(10):54.

\bibitem[{Yang et~al.(2024)Yang, Jin, Zhu, Wang, Álvarez, Wan, Hou, and Lu}]{paper_survey_metrics}
Yifan Yang, Qiao Jin, Qingqing Zhu, Zhizheng Wang, Francisco~Erramuspe Álvarez, Nicholas Wan, Benjamin Hou, and Zhiyong Lu. 2024.
\newblock \href {https://arxiv.org/abs/2410.18460} {Beyond multiple-choice accuracy: Real-world challenges of implementing large language models in healthcare}.
\newblock \emph{Preprint}, arXiv:2410.18460.

\bibitem[{Yu et~al.(2025)Yu, Jin, Shu, Zhang, Fan, Hua, Zhu, Meng, Wang, Du, and Zhang}]{health-llm}
Qinkai Yu, Mingyu Jin, Dong Shu, Chong Zhang, Lizhou Fan, Wenyue Hua, Suiyuan Zhu, Yanda Meng, Zhenting Wang, Mengnan Du, and Yongfeng Zhang. 2025.
\newblock \href {https://arxiv.org/abs/2402.00746} {Health-llm: Personalized retrieval-augmented disease prediction system}.
\newblock \emph{Preprint}, arXiv:2402.00746.

\bibitem[{Zhou et~al.(2024{\natexlab{a}})Zhou, Liu, Gu, Zou, Huang, Wu, Li, Chen, Zhou, Liu, Hua, Mao, You, Wu, Zheng, Clifton, Li, Luo, and Clifton}]{paper_survey_basic}
Hongjian Zhou, Fenglin Liu, Boyang Gu, Xinyu Zou, Jinfa Huang, Jinge Wu, Yiru Li, Sam~S. Chen, Peilin Zhou, Junling Liu, Yining Hua, Chengfeng Mao, Chenyu You, Xian Wu, Yefeng Zheng, Lei Clifton, Zheng Li, Jiebo Luo, and David~A. Clifton. 2024{\natexlab{a}}.
\newblock \href {https://arxiv.org/abs/2311.05112} {A survey of large language models in medicine: Progress, application, and challenge}.
\newblock \emph{Preprint}, arXiv:2311.05112.

\bibitem[{Zhou et~al.(2024{\natexlab{b}})Zhou, Lin, Ding, Wang, Melton, Zou, and Zhang}]{dual_inf}
Shuang Zhou, Mingquan Lin, Sirui Ding, Jiashuo Wang, Genevieve~B. Melton, James Zou, and Rui Zhang. 2024{\natexlab{b}}.
\newblock \href {https://arxiv.org/abs/2407.07330} {Interpretable differential diagnosis with dual-inference large language models}.
\newblock \emph{Preprint}, arXiv:2407.07330.

\end{thebibliography}

\appendix

\clearpage
\newpage

\section{Related work}\label{appx:related_works}

To describe the current state of medical dialogue systems, we first examine their evolution, highlighting how they have progressed from rule-based systems to modern LLM-powered AI. Then, we identify key challenges that must be addressed to develop robust and clinically reliable AI-driven consultations.

\subsection{Evolution of Medical Dialogue Systems}
Medical dialogue systems have evolved through three major stages:

\begin{enumerate}
    \item \textbf{Rule-Based Expert Systems} – Early models such as Mycin \cite{mycin} and DXplain\cite{DXplain} used manually encoded medical knowledge and decision trees. While interpretable, they lacked adaptability and could not handle open-ended patient input.
    
    \item \textbf{Task-Specific AI Models} – Machine learning-based systems such as Babylon Health \cite{babylong} and Ada Health \cite{adahealth} introduced data-driven triage and symptom assessment, but they remained single-turn and failed to engage in dynamic diagnostic hypotheses transparency.
    
    \item \textbf{LLM-Powered Conversational Agents} – Recent systems, such as AMIE \cite{amie_model}, DISC-MedLLM \cite{DISC_model}, CoD \cite{CoD_model} and MEDAGENTS \cite{MedAgents}, integrate structured multi-turn dialogue processing and medical reasoning, improving diagnostic hypotheses accuracy and interaction depth.
\end{enumerate}

Despite these advancements, medical dialogue systems still struggle with fundamental issues related to structured reasoning, factual reliability, and real-time decision-making.

\subsection{Challenges in Medical AI Conversations}

Medical dialogue systems have made significant progress, but their deployment in real-world clinical settings remains challenging. Despite leveraging LLMs for more natural and interactive patient communication, these systems exhibit critical limitations that hinder their reliability and effectiveness. The primary challenges include structured reasoning deficiencies, factual inaccuracies, and failure to prioritize critical cases.

\subsubsection{Insufficient Transparency in Multi-Turn Diagnostic Dialogue}
One of the primary shortcomings of current medical dialogue systems is their inability to conduct structured, multi-turn reasoning. Many models generate fragmented, inconsistent, or overly simplistic responses, failing to follow a logical sequence similar to a physician’s diagnostic hypotheses inquiry.

Several approaches have been proposed to address the challenges of structured diagnostic hypotheses reasoning in medical dialogue systems. Chain of Diagnosis (CoD) introduces a framework designed to mimic the stepwise reasoning of physicians, systematically guiding the diagnostic hypotheses process and reducing the risk of premature conclusions~\cite{CoD_model}. Another notable system, AMIE, leverages a self-play training mechanism within a simulated environment, enabling the model to iteratively refine both its questioning strategies and decision-making capabilities across a variety of medical scenarios~\cite{amie_model}. Similarly, DISC-MedLLM enhances the depth of structured questioning by integrating medical knowledge graphs and training on real-world, multi-turn medical dialogues, ensuring that the system can navigate complex patient interactions with greater precision~\cite{DISC_model}.

Despite these advancements, challenges persist. Models struggle with handling ambiguous patient responses, dynamically adjusting diagnostic hypotheses strategies, and effectively incorporating context across long multi-turn conversations.

\subsubsection{Hallucinations and Misinformation}
A significant limitation of LLM-based medical dialogue systems is their tendency to generate factually incorrect or misleading information, known as hallucinations. These inaccuracies can lead to diagnostic hypotheses errors and compromise patient safety.

Several strategies have been proposed to mitigate the challenges associated with medical dialogue systems. Health-LLM utilizes a retrieval-augmented generation (RAG) mechanism to ground AI-generated responses in trusted medical sources, ensuring real-time verification and reducing the risk of hallucinations~\cite{health-llm}. In practice, retrieval and grounding can be supported by domain corpora that link clinical text to controlled terminologies (e.g., UMLS/ICD), such as NEREL-BIO, RuCCoN, and RuCCoD \citep{NERELBIO,loukachevitch2024biomedical,Nesterov2022239,nesterov2025ruccod}. Similarly, MEDAGENTS implements a multi-agent framework in which AI-based "experts" collaboratively cross-validate each other's responses, thereby minimizing incorrect outputs and enhancing overall reliability~\cite{MedAgents}. Another approach, DISC-MedLLM, combines structured multi-turn dialogues with the integration of medical knowledge graphs to provide detailed explanations for each diagnostic step. This design not only improves the transparency of the reasoning process but also facilitates validation by enabling clinicians to trace how conclusions are reached~\cite{DISC_model}.

Nonetheless, hallucinations remain prevalent in complex diagnostic hypotheses cases, particularly those involving rare diseases, overlapping symptoms, or mental health conditions. Ensuring real-time fact-checking and clinician-in-the-loop validation remains a pressing challenge.

\subsubsection{Failure to Recognize Critical Conditions}
Medical AI systems often fail to detect and escalate high-risk conditions requiring urgent medical attention. This limitation reduces their effectiveness in emergency triage and high-stakes diagnostic hypotheses settings.

Several approaches have been developed to improve the recognition of critical conditions in medical dialogue systems. Dual-Inf enhances diagnostic hypotheses reliability by employing bidirectional inference, allowing the model to cross-reference symptoms with potential diagnostic hypotheses to detect severe conditions more accurately~\cite{dual_inf}. Similarly, AMIE integrates an uncertainty-aware decision-making mechanism, which prompts the model to request additional clarifications whenever its confidence in a diagnostic hypotheses is low, thereby reducing the likelihood of incorrect or premature conclusions~\cite{amie_model}. Another approach, Polaris, leverages specialized risk assessment models to identify high-risk cases and escalate them for human intervention when necessary, ensuring that critical conditions receive appropriate attention~\cite{polaris}.

However, LLM-based medical systems still struggle with distinguishing time-sensitive symptoms from non-urgent ones, balancing AI autonomy with human intervention, and responding efficiently to emergency cases.

\section{Methods}
\subsection{Proposed Solution to Challenges}
\begin{table*}[h!]
\centering
\begin{adjustbox}{max width=\textwidth}
\begin{tabular}{@{}ll|ccc|ccc@{}}
\toprule
\textbf{Category}            & \textbf{System}         & \textbf{Transparency} & \textbf{Safety} & \textbf{Critical} & \textbf{Real-time} & \textbf{Adaptability} & \textbf{Scalability} \\ \midrule
\textbf{Rule-based}          & MyCin\cite{mycin}                   & \textbf{+-}         & \textbf{+}      & \textbf{+-}       & \textbf{+}         & \textbf{-}            & \textbf{-}           \\ 
                             & Dxplain \cite{DXplain} & \textbf{+-}         & \textbf{+}      & \textbf{+-}       & \textbf{+}         & \textbf{-}            & \textbf{-}           \\ \midrule
\textbf{Task-specified}      & Babylon Health \cite{babylong}          & \textbf{-}        & \textbf{-}      & \textbf{+-}       & \textbf{+}         & \textbf{-}            & \textbf{-}           \\ 
                             & Ada Health \cite{adahealth}              & \textbf{-}        & \textbf{-}      & \textbf{+-}       & \textbf{+}         & \textbf{-}            & \textbf{-}           \\ \midrule
\textbf{LLM-based}           & DISC-MedLLM \cite{DISC_model}            & \textbf{+}         & \textbf{+-}     & \textbf{+}        & \textbf{+}         & \textbf{+-}           & \textbf{+-}          \\ 
                             & MEDAGENTS \cite{MedAgents}              & \textbf{+-}        & \textbf{+}      & \textbf{+}        & \textbf{+-}        & \textbf{+-}           & \textbf{-}           \\ 
                             & AMIE \cite{amie_model}                   & \textbf{+}         & \textbf{+}      & \textbf{+}        & \textbf{+-}        & \textbf{+-}           & \textbf{+-}          \\ 
                             & Health-LLM \cite{health-llm} & \textbf{+-}        & \textbf{+-}     & \textbf{+}        & \textbf{+-}        & \textbf{-}            & \textbf{-}           \\ 
                             & Polaris \cite{polaris}                & \textbf{+}         & \textbf{+}      & \textbf{+}        & \textbf{+}         & \textbf{+-}           & \textbf{+-}          \\ 
                             & Dual-Inf \cite{dual_inf}               & \textbf{+}         & \textbf{+}      & \textbf{+}        & \textbf{+}         & \textbf{+-}           & \textbf{-}           \\ 
                           
~                & \sysname{} (\textbf{ours})                 & \textbf{+}         & \textbf{+}      & \textbf{+}        & \textbf{+}         & \textbf{+}            & \textbf{+}           \\ \bottomrule
\end{tabular}
\end{adjustbox}
\caption{
\small
Qualitative capability matrix. Each ‘+’ corresponds to a documented, peer-reviewed evaluation; ‘±’ indicates partial support or missing evidence.}
\label{tab:medical_ai_comparison}

\end{table*}
Our proposed system, \sysname{}, offers a comprehensive solution to the challenges faced by existing medical dialogue systems, as summarized in Table~\ref{tab:medical_ai_comparison}. By integrating Finite State Machines (FSMs) with Large Language Models (LLMs), \sysname{} outperforms other approaches across key dimensions: structured reasoning, safety, critical condition recognition, real-time readiness, adaptability, and scalability.

\textbf{Transparency}: \sysname{} ensures consistent and structured multi-turn diagnostic hypotheses reasoning through FSMs, mimicking the systematic approach of experienced clinicians. In contrast, while systems like DISC-MedLLM and AMIE exhibit strong reasoning capabilities, they lack the robustness of FSM-driven workflows, leading to fragmented or inconsistent interactions in complex cases.

\textbf{Safety}: One of the most pressing challenges in medical AI is hallucination and misinformation. \sysname{} addresses this through a Retrieval-Augmented Generation (RAG) approach, grounding all outputs in trusted medical knowledge bases. Systems like Polaris and Dual-Inf attempt similar mitigation strategies but fall short of the transparency and reliability offered by \sysname{}.

\textbf{Critical Condition Recognition}: The probabilistic risk triage model in \sysname{} ensures real-time identification and escalation of high-risk cases, outperforming solutions like MEDAGENTS, which lack the precision and adaptability needed for critical diagnostic hypotheses.

\textbf{Real-time Readiness}: \sysname{} is designed for seamless integration into clinical environments, leveraging modular microservices for real-time deployment. Task-specific systems such as Babylon Health and Ada Health, while functional, are limited to specific use cases and lack the flexibility of \sysname{}.

\textbf{Adaptability}: Unlike other systems, \sysname{} dynamically adjusts to diverse scenarios, including rare diseases and ambiguous patient inputs, due to its hybrid architecture. LLM-based systems like Health-LLM and AMIE struggle in these areas due to limited context-awareness and reliance on static models.

\textbf{Scalability}: The modular design of \sysname{} enables effortless scaling across various clinical infrastructures, making it suitable for both large hospital networks and smaller healthcare providers. Existing solutions, including DISC-MedLLM and Polaris, lack this level of deployment flexibility.

In summary, \sysname{} demonstrates unmatched performance across all key criteria, bridging the gap between cutting-edge AI research and practical healthcare applications. By combining structured reasoning, advanced safety measures, and real-time adaptability, \sysname{} sets a new standard for medical dialogue systems.

\section{Dialogue examples}\label{appx:dialogue_examples}

\begin{table*}[h!]
\centering
\begin{adjustbox}{max width=\textwidth}

\begin{tabular}{|p{1.3cm}|p{4.5cm}|p{3.5cm}|p{3.5cm}|p{3.5cm}|} \hline
 & \textbf{Transparency} & \textbf{Critical} & \textbf{Safety} & \textbf{Adaptability} \\ \hline

% Вторая строка
\textbf{Dialogue example} &
\begin{minipage}[t]{\linewidth}
\textbf{Doc:} What’s bothering you?\\
\textbf{Pt:} I have a headache.\\
\textbf{Doc:} Where exactly is the pain located?\\
\textbf{Pt:} The back of my head.\\
\textbf{Doc:} Are you experiencing any other symptoms, such as nausea, vomiting?\\
\textbf{Pt:} No.\\
\textbf{Doc:} How intense is the pain?\\
\textbf{Pt:} 5 out of 10.\\
\textbf{Doc:} \\
Cervicogenic headache – General practitioner. Pain in the back of the head may be related to issues in the cervical spine.\\
Cervical osteochondrosis – Neurologist. You have a neck condition that could be causing the headache in the back of your head.\\
\textbf{Doc:} Is everything clear? Feel free to ask questions!
\end{minipage} 
&
\begin{minipage}[t]{\linewidth}
\textbf{Doc:} What is bothering you?\\
\textbf{Pt:} High blood pressure.\\
\textbf{Doc:} Do you experience chest pain or an increased heart rate?\\
\textbf{Pt:} Yes.\\
\textbf{Doc:} Do you have any vision problems?\\
\textbf{Pt:} Yes.\\
\textbf{Doc:} \\
Your condition could be close to critical!\\
Call 103 immediately.\\
Wait for the response, do not hang up!\\
Briefly explain what happened.\\
\end{minipage} 
&
\begin{minipage}[t]{\linewidth}
\textbf{Doc:} What’s bothering you?\\
\textbf{Pt:} Who are you?\\
\textbf{Doc:} I am a neural network model that can assist with medical questions. How can I help you?\\
\textbf{Pt:} Help me with algorithms in Python.\\
\textbf{Doc:} I’m sorry, I didn’t understand your response. Could you please provide more specific information or rephrase your message?\\
\textbf{Pt:} What can you help me with?\\
\textbf{Doc:} I can provide consultations on health and medical issues. Please tell me more about your symptoms or concerns so that I can offer appropriate assistance.\\
\end{minipage} 
&
\begin{minipage}[t]{\linewidth}
\textbf{Doc:} What’s bothering you?\\
\textbf{Pt:} I’ve been smoking for 5 years, half a pack a day. I’ve started coughing up very thick phlegm. How can I check my lungs? Could you provide a list of examinations and doctors I need to see?\\
\textbf{Doc:} You need to see a general practitioner (therapist) or a pulmonologist first. The doctor will order a chest X-ray or a CT scan of the lungs to rule out pathological changes.\\
\end{minipage} \\ \hline
\hline
\textbf{Related services metrics} &
Information collector: $P=84\,\%$, $f_{\text{rep}}=0$\par
Readiness estimator: $\text{MAPE}=22\,\%$\par
Medical-specialty selector: $R@3=96\,\%$ &
Emergency detector: $F_{1}=59\,\%$, FPR $=2\,\%$ &
Moderator: $F_{1}=95\,\%$ &
Question detector: $F_{1}=94\,\%$\\
\hline
\end{tabular}
\end{adjustbox}
\caption{\label{tab:example_full}\sysname{} dialogue examples and full-study results with relevant metrics.}
%\vspace{-0.7cm}
\end{table*}

\section{Detailed Component-Level Results}
\label{appx:detailed_results}

\subsection{Moderator}
The moderator was evaluated on a dataset of 10,000 messages, achieving an $F_1$-score of 0.95. The system demonstrated high accuracy in detecting prohibited topics, with a false positive rate of less than 1.5\%.

\subsection{Emergency Detector}
We trained the emergency detection module on 2,114 medical dialogues and evaluated it on a held-out set of 682 chats, each independently annotated by three licensed physicians. At the operating threshold t=0.11, selected in alignment with stakeholder requirements for high reliability, the system achieved a precision of 0.72, an F$_1$-score of 0.59, and a false positive rate (FPR) of just 0.02. This threshold was deliberately optimized to minimize false alarms in real-world deployment, ensuring the system maintains a high level of trust and stability in routine clinical use. While the current configuration favors specificity, the model architecture allows for straightforward adjustment of this threshold to support more recall-oriented use cases, such as emergency triage scenarios. This adaptability makes the module suitable for diverse clinical workflows, and its consistent performance across evaluation settings provides a strong foundation for safe and effective integration.

Beyond the primary emergency detector, the safety stack includes:
\begin{itemize}
    \item A domain-specific moderator that filters unsafe topics at input and output stages.
    \item Deterministic FSM gates that prevent transition to the “Diagnostic Hypothesis \& Routing” state unless readiness is verified, holding low-confidence dialogues for human handoff.
    \item A universal red-flag questionnaire presented to all users as a final safety net. 
\end{itemize}

\subsection{Readiness Estimator}
The readiness estimator was trained on a corpus of 2,500 medical dialogues, with each message labeled as part of the anamnesis collection or diagnostic hypotheses stage. The model used TF-IDF and contextual embeddings for feature extraction, achieving an $F_1$-score of 0.78. The mean absolute percentage error (MAPE) for predicting the duration of the anamnesis collection phase was 22\%.

\subsection{Question Detector}
A total of 9,077 user messages were annotated to train the question detector. The classifier, based on logistic regression and contextual embeddings, achieved a recall of 0.87 for the positive class, a precision of 0.99 for the negative class, and a macro-averaged $F_1$-score of 0.94. These results demonstrate high accuracy in identifying clarifying questions.

\subsection{Information Collector}
The information collector module was tested on 1,000 dialogues, achieving a precision of 0.84 for question relevance. The average response time was 5 seconds, with 20\% of cases utilizing cached questions from the historical dialogue database.

\subsection{Medical Specialty Selector}
Medical specialty selection system input $\mathfrak{W} = \{W_1, ..., W_n\}, n \in \mathbb{N}$ is a set of $n$ medical documents $W$. $W$ is a text string that contains a chat between a patient and a doctor. Medical specialty selection system output $\mathfrak{D} =[\![ D_1, ..., D_n]\!] , n \in \mathbb{N}$ is a multiset of $n$ multisets $D = [\![ d_1, ..., d_k ]\!], k \in \mathbb{N}$, where $d$ is a text string that contains a doctor specialty. Expectations of system performance will be shaped by the set of experts $\mathfrak{E} = \{E_1, ..., E_z\}, z \in \mathbb{N}$, where expert is a function $E:\mathfrak{W} \mapsto \mathfrak{D}$. The system itself is a function $A:\mathfrak{W} \mapsto \mathfrak{D}$  which is not in a set of experts: $\chi_{\mathfrak{E}}(A)= \emptyset$. Then we can define the pairwise precision $P$ and recall $R$ between the algorithm and the expert as 
\begin{equation}
    P_{AE}@k = \frac{\sum_{i=1}^{n}\mu(D^A_i, D^E_i)}{nk^2}    
\end{equation}
\begin{equation}
    R_{AE}@k = \frac{\sum_{i=1}^{n}\chi(D^A_i, D^E_i)}{n}  
\end{equation}
where $D_i^A = A(W_i)$, $D_i^E = E(W_i)$,  $\mu$ is a multiplicity function, and $\chi$ is a characteristic function. We consider the accuracy of the model's answer by the complete match of its answer and the expert's answer. Thus, we can calculate the multiplicity $\mu$ and the characteristic $\chi$ functions in case of $|D^A| = |D^E| = k$:
\begin{equation}
   \mu(D^A, D^E) = \sum_{p=1}^{k}\sum_{q=1}^{k} 
    \begin{dcases}
        1 & d^A_p = d^E_q \\
        0 & d^A_p \neq d^E_q
    \end{dcases} 
\end{equation}

\begin{equation}
    \chi(D^A, D^E) = 
    \begin{dcases}
        1 & \mu(D^A, D^E) > 0 \\
        0 &   \mu(D^A, D^E) = 0
    \end{dcases}    
\end{equation}
The raw dataset is a collection of $n=360$ chats. The first half of the dataset consists of chats from two doctors played out a pre-determined scenario. One of them knew the disease and their symptoms and was the patient. The other tried to define the disease by asking questions as the doctor. The second half of the dataset consists of the chats between the actor playing the role of a patient and \sysname{} as the doctor. Ground truth labels for the dataset were marked by $z=7$ experts. Two of them are licensed doctors and other 8 are resident physicians. Each expert marked up all chats with up to $k_{max}=5$ answers.  The quality metrics presented in Table \ref{tab:PQ} are more than sufficient for product use.  It is also interesting that the model ranks the answers in order of importance very well. The share of the first answer among the total number of correct answers is 80\%, the share of the first two answers is 95\%.

\begin{table}[htbp]
    \centering
        \caption{\label{tab:PQ} Pairwise quality of the medical specialty selector compared to all experts. The number of specialists in the answer is equal to $k$.}\begin{tabular}{|c|c|c|}
         \hline
               \textbf{k} & \textbf{Precision@k} & \textbf{Recall@k} \\ \hline
               1 & 77$\pm$0.3\%&  77$\pm$0.3\%\\ \hline
               2 & 71$\pm$0.2\%&  92$\pm$0.4\%\\ \hline
               3 & 68$\pm$0.3\%&  96$\pm$0.4\%\\\hline
    \end{tabular}
\end{table}

\subsection{Language Model Details}
All LLM-based components (answer generation, diagnostic hypothesis drafting, and explanation synthesis) are powered by \llmname{}. 
We used the production configuration provided by the vendor; prompts and safety constraints are listed in Appendix~\ref{appx:detailed_results}. 
Where relevant, we apply deterministic decoding for safety-critical paths and temperature-limited sampling for open-dialogue responses.

\section{Formal Description of the FSM}\label{appx:fsm_description}

Let the Finite State Machine be defined as \\ a tuple $M = (Q, \Sigma, \Omega, C, T, \delta, \lambda, q_0, q_d, q_{ca}, F)$ where:
\begin{itemize}
    \item $Q$ is a finite set of states.
    \item $W$ is the input alphabet, defined as the set of all possible strings representing user messages and previous chat history.
    \item $\Omega$ is the output alphabet, defined as the set of all possible text messages generated by the system.
    \item $C$ is the set of all conditions for transitions.
    \item $T$ is the set of all transitions.
    \item $\delta: Q \times W \rightarrow Q$ is the transition function.
    \item $\lambda: Q \times W \rightarrow \Omega$ is the output function.
    \item $q_0 \in Q$ is the initial state.
    \item $q_d \in Q$ is the default state.
    \item $q_{ca} \in Q$ is the state with the correct answer.
    \item $F \subseteq Q$ is the set of final states.
    \item $N\_ATTEMPTS$ is the maximum number of attempts allowed in the internal cycle.
\end{itemize}

\subsection{Set of Conditions $C$}
The set of conditions $C$ is defined as:
\[ C = \{ c_1, c_2, \ldots, c_m \} \]
where each $c_i$ is a condition that checks if the input $w \in W$ satisfies certain criteria.

\subsection{Set of Transitions $T$}
The set of transitions $T$ is defined as:
\[ T = \{ t_1, t_2, \ldots, t_n \} \]
where each $t_j$ is a transition defined by:
\[ t_j = (q_j, c_j, q'_j) \]
where:
\begin{itemize}
    \item $q_j \in Q$ is the source state.
    \item $c_j \subseteq C$ is the set of conditions for the transition.
    \item $q'_j \in Q$ is the target state.
\end{itemize}

Each transition $t_j$ has an associated function $is\_triggered(w)$ that determines if the transition is triggered by the input $w \in W$. This function is defined as:
\[ t_j.is\_triggered(w) = \bigwedge_{i=1}^{|c_j|} c_i(w) \]

\subsection{Transition Function}\label{appx:fsm_transition}
The transition function $\delta$ is defined as:
\[
\delta(q,w)=
\left\{
\begin{array}{ll}
q'  & \begin{aligned}[t]
         &\text{if }\exists\,t\in T:\\
         &\quad t.\operatorname{is\_triggered}(w)=\text{true}
       \end{aligned}\\[4pt]
q_d & \text{otherwise}
\end{array}
\right.
\]

where:
\begin{itemize}
    \item $T$ is the set of all transitions.
    \item $t.is\_triggered(w)$ checks if the input $w \in W^*$ received from the user satisfies the conditions of transition $t$.
\end{itemize}

\subsection{Output Function}\label{appx:fsm_outputs}
The output function $\lambda$ is defined as:
\[ \lambda: Q \times W \rightarrow \Omega \]
\[ \lambda(q, w) = action_q(w) \]
where $action_q$ is the function associated with state $q$ that generates the system response based on the input $w \in W^*$ received from the user.

\subsection{External Cycle}
The external cycle represents the system-user interaction sequence:
\[ (q_i, w_i) \vdash (q_{i+1}, w_{i+1}, r_i) \]
where:
\begin{itemize}
    \item $q_i \in Q$ is the current state.
    \item $w_i \in W^*$ is the input string received from the user.
    \item $r_i \in \Omega$ is the system response.
\end{itemize}

The external cycle continues until the user interrupts it.

\subsection{Internal Cycle}
The internal cycle is defined as:
\[ (q, w) \vdash^* (q', r) \]
where:
\begin{itemize}
    \item $\vdash^*$ is the reflexive transitive closure of $\vdash$.
    \item $q'$ is the state reached after at most $N\_ATTEMPTS$ steps.
    \item $r \in \Omega$ is the system response.
\end{itemize}

The internal cycle represents a single step in the external cycle, where the system generates a response based on the user input.

\subsection{Termination Conditions}
The internal cycle terminates when one of the following conditions is met:
\[ q \in F \lor |inner\_states| = N\_ATTEMPTS \lor q = q_{ca} \]

\subsection{Output}
The output of the internal cycle is defined as a tuple:
\[ (r, q) \in \Omega \times Q \]
where $r$ is the system response and $q$ is the next state for the external cycle.

\begin{algorithm}
\caption{FSM Algorithm}
\begin{algorithmic}[1]
    \State $q \gets q_0$ \Comment{Set the initial state}
    \While{not user\_interrupted} \Comment{External Cycle}
        \State $w \gets user\_input$ \Comment{Get the user input}
        \State $inner\_states \gets \emptyset$ \Comment{Initialize the inner states list}
        
        \While{$q \notin F \land |inner\_states| < N\_ATTEMPTS \land q \neq q_{ca}$} \Comment{Internal Cycle}
            \State $q' \gets \delta(q, w)$ \Comment{Apply the transition function}
            \State $r \gets \lambda(q', w)$ \Comment{Generate the output}
            \State $inner\_states \gets inner\_states \cup \{(q', r)\}$ \Comment{Add the state-output pair to inner states}
            \State $q \gets q'$ \Comment{Update the current state}
        \EndWhile
        
        \State \Return $r$ \Comment{Return the system response}
    \EndWhile
\end{algorithmic}
\end{algorithm}

\section{Information Collector Diagram}\label{appx:InfCollDiag}
\begin{center}
    \includegraphics[width=\columnwidth]{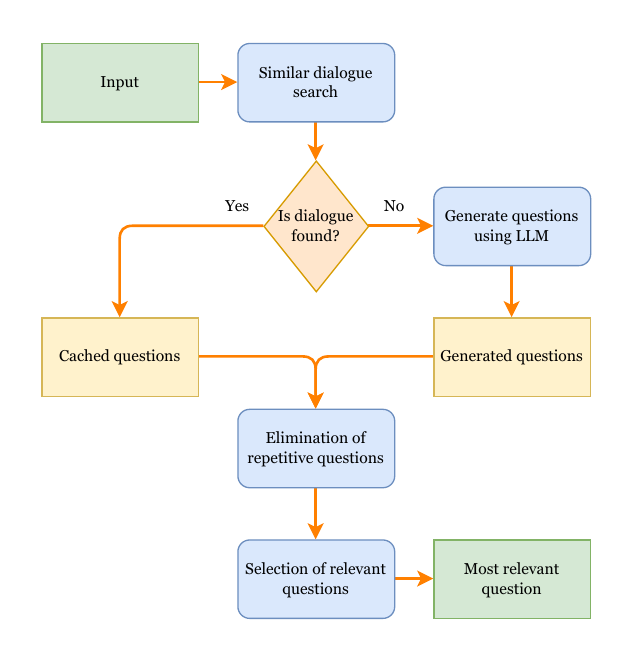}
\end{center}

\section{Medical Specialty Selector Diagram}\label{appx:SpecSelDiagram}
\begin{center}
    \includegraphics[width=\columnwidth]{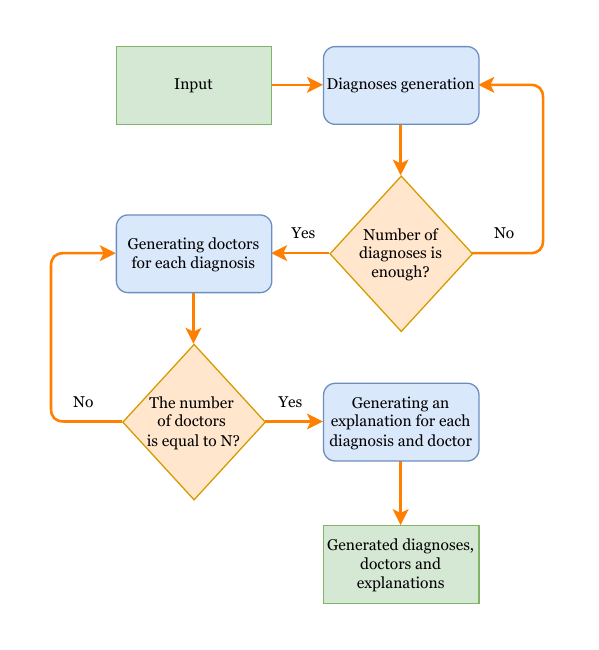}
\end{center}

\section{Gender distribution}\label{appx:gender_dist}
\begin{center}
    \centering
    \includegraphics[width=\columnwidth]{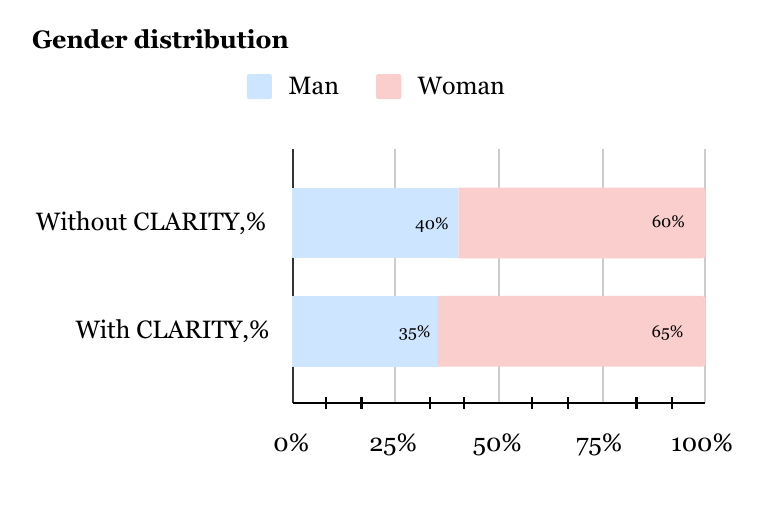}
\end{center}

\section{Age groups distribution}\label{appx:age_dist}
\begin{center}
    \centering
    \includegraphics[width=\columnwidth]{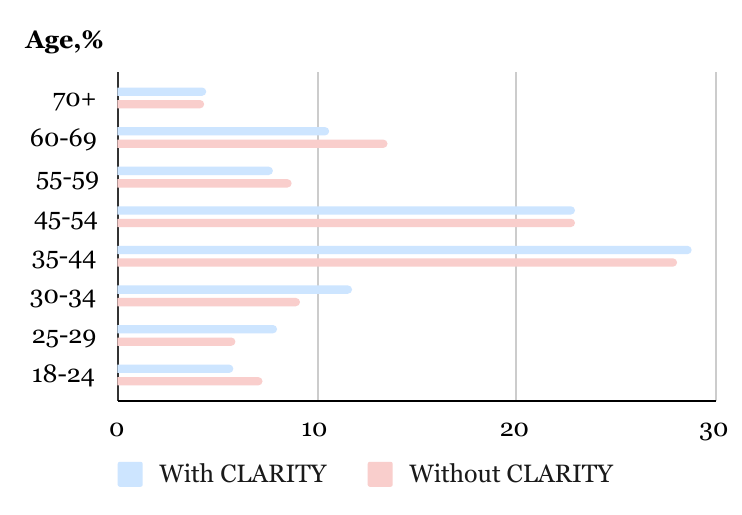}
\end{center}

\section{User survey results: interaction experience.}\label{appx:commun_friendly}
\begin{center}
    \centering
    \includegraphics[width=1\linewidth]{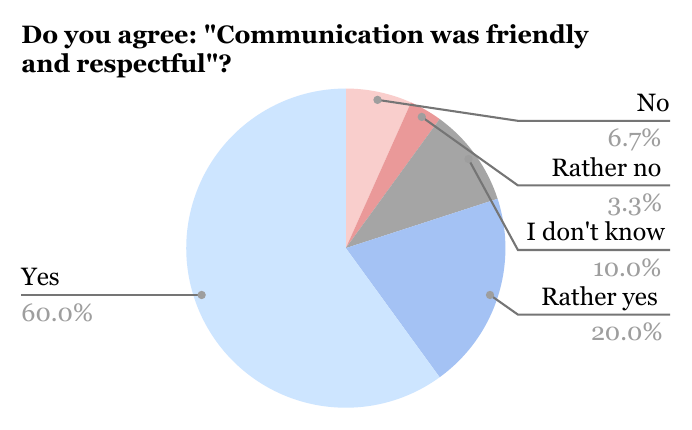}
\end{center}

\section{Statistics on the recommended specialists in the expert-validated set of dialogues.}\label{appx:recom_spec}
\begin{center}
   \centering
   \includegraphics[width=1\linewidth]{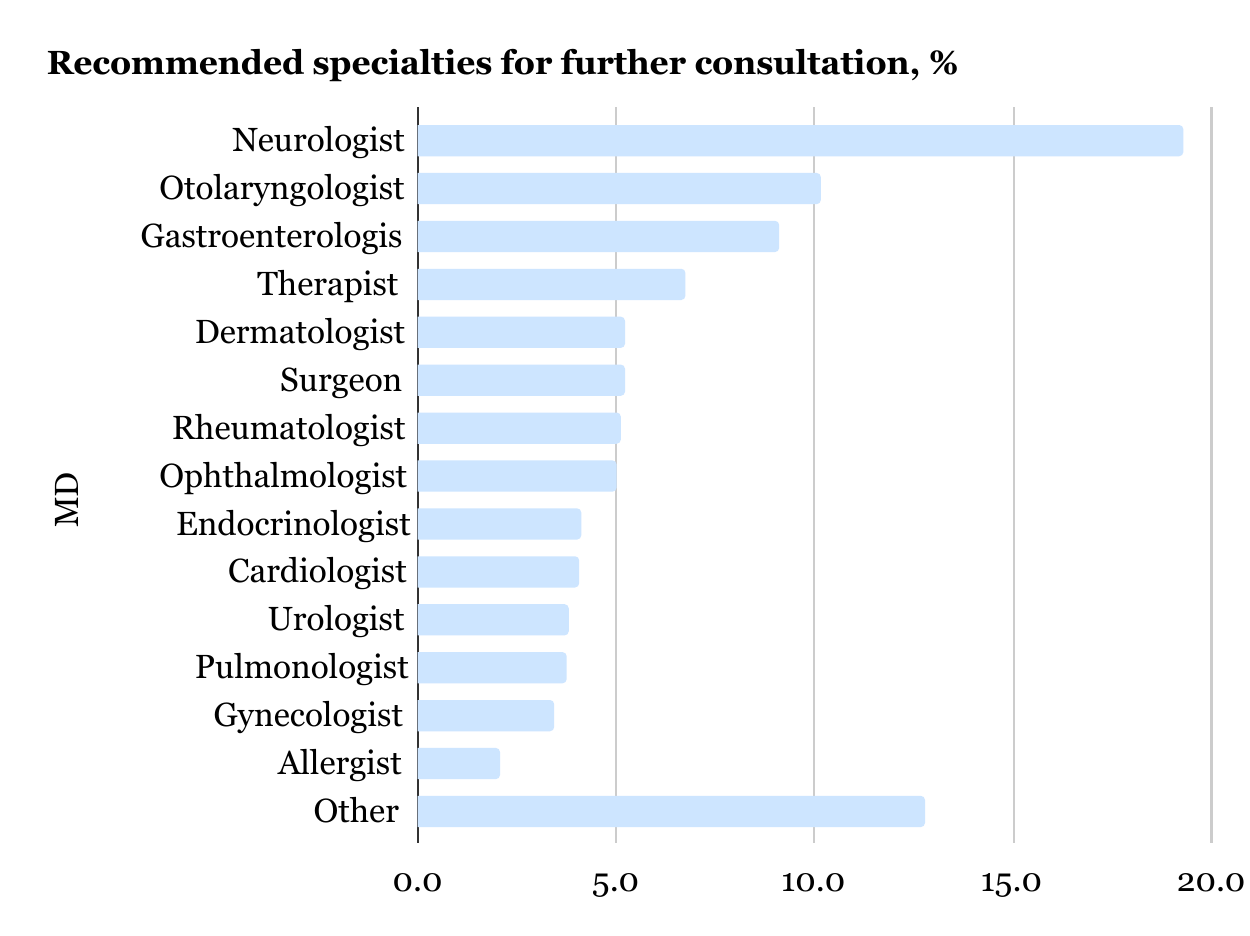}
\end{center}

\section{User input description}

The user interface of the \sysname{} system exposes the \verb|/v3/request| request URL for queries of various medical scenarios.

The user request body has the following contents
\begin{verbatim}
{
  "Text": "User message",
  "OuterContext": {
    "Sex": true,
    "Age": 21,
    "UserId": "UserId",
    "SessionId": "SessionId",
    "ClientId": "ClientId",
  }
},
\end{verbatim}
where the \verb+"Text"+ field contains the input user message and the \verb+"OuterContext"+ field contains all the necessary information about the user. Inside the \verb+"OuterContext"+ the \verb+"Sex"+ and \verb+"Age"+ fields denote the user's age and sex, respectively. The \verb+"UserId"+, \verb+"SessionId"+ and \verb+"ClientId"+ fields together form a unique dialogue identifier, which is used across the \sysname{} components for dialogue context propagation.

\section{System output description}

In general, the \sysname{} system output has the following contents
\begin{verbatim}
{
  "Text": "System output message",
  "Results": [...]
},
\end{verbatim}
where the \verb+"Text"+ field contains the system output message and the \verb+"Results"+ field contains additional information on the dialogue context. The contents of the \verb+"Results"+ field depend on the dialogue state. For any dialogue state except for the \textit{Diagnostic hypotheses state} the \verb+"Results"+ field is empty. For the \textit{Diagnosis state} dialog state the \verb+"Results"+ array contains three elements each having the following structure.
\begin{verbatim}
{
  "Diagnosis": "One of three possible diagnoses",
  "Doctor": "Recommended medical specialty",
  "Description": "Diagnosis description"
},
\end{verbatim}
where the \verb|"Diagnosis"| field contains the name of the diagnosis, the \verb|"Doctor"| field contains the suggested medical specialty for further consultation and the \verb|"Description"| field contains text which clarifies the diagnosis name.

\end{document}